\documentclass[pre,twocolumn,superscriptaddress,floatfix]{revtex4-1}  

\usepackage[paperwidth=210mm,paperheight=297mm,centering,hmargin=2cm,vmargin=2.5cm]{geometry}

\usepackage[pdftex]{graphicx}
\usepackage{amsmath,amsfonts,amssymb}
\usepackage{natbib}
\usepackage{MnSymbol}
\usepackage{csquotes}

\usepackage{hyperref}
\usepackage{xcolor} 
\usepackage{array}
\usepackage{pgfplots}
\usepackage{graphicx}
\usepackage[left,modulo,pagewise]{lineno}
\usepackage{dsfont}

\definecolor{bluemoi}{rgb}{0.25,0.50 ,0.75} 

\hypersetup{
  bookmarksopen=true,
  pdftitle=" ",
  pdfauthor=" ", 
  pdfsubject=" ", 
  pdftoolbar=true, 
  pdfmenubar=true, 
  pdfhighlight=/O, 
  colorlinks=true, 
  pdfpagemode=UseNone, 
  pdfpagelayout=SinglePage, 
  pdffitwindow=true, 
  linkcolor=bluemoi, 
  citecolor=bluemoi, 
  urlcolor=bluemoi 
}

\makeatletter
\renewcommand{\figurename}{\sf \textbf{Figure}}
\renewcommand{\thefigure}{\arabic{figure}}
\renewcommand{\fnum@figure}{\sf\textbf{\figurename}~\textbf{\thefigure}}
\renewcommand{\tablename}{\sf\textbf{Table}}
\renewcommand{\thetable}{\arabic{table}}
\renewcommand{\fnum@table}{\sf\textbf{\tablename}~\textbf{\thetable}}
\makeatother

\begin{document}

\title{Generating OWA weights using truncated distributions} 

\author{Maxime Lenormand}
\thanks{Corresponding author: maxime.lenormand@irstea.fr}
\affiliation{Irstea, UMR TETIS, 500 rue JF Breton, FR-34093 Montpellier, France}

\begin{abstract} 
Ordered weighted averaging (OWA) operators have been widely used in decision making these past few years. An important issue facing the OWA operators' users is the determination of the OWA weights. This paper introduces an OWA determination method based on truncated distributions that enables intuitive generation of OWA weights according to a certain level of risk and trade-off. These two dimensions are represented by the two first moments of the truncated distribution. We illustrate our approach with the well-know normal distribution and the definition of a continuous parabolic decision-strategy space. We finally study the impact of the number of criteria on the results.
\end{abstract}

\maketitle

\section*{Introduction}

The ordered weighted averaging (OWA) operator was first introduced by Yager in 1998 \citep{Yager1988} to address the problem of aggregating a set of criteria functions in order to form an overall decision function. This seminal paper have shed some light on the recurrent problem of aggregation shared by many disciplines and used in a wide range of applications \citep{Yager2011}. In particular, OWA operators have been extensively used in Multiple Criteria Decision Analysis (MCDA) to provide support in complex decision-making situations. OWA operators have notably been applied in GIS environments for the production of land-use suitability maps \citep{Malczewski2006} with applications in land-use planning and management such as landslide susceptibility mapping \citep{Feizizadeh2014}, mapping of wilderness \citep{Comber2010}, ecological capability assessment \citep{Ferretti2013}, disease susceptibility mapping \citep{Dong2016} or soil fertility evaluation \citep{Mokarram2017}. OWA operators are indeed particularly interesting in decision analysis, because they offer the possibility to incorporate a certain level of risk and trade-off in decision-making processes \citep{Jiang2000,Drobne2009}. These two concepts are inextricably connected to the distribution of OWA weights, the form of which implies the satisfaction of one or several criteria, reflecting a certain risk inclination/aversion and willingness to compromise. 

An important issue regarding the manipulation of OWA operators is to determine the weights. The result of aggregation is indeed highly dependent of the vector of OWA weights used to aggregate the criteria. OWA weights generation is an active research topic and numerous approaches have been proposed during the past few years \citep{Liu2011}. First, we need to make a distinction between OWA weights determination methods based on external information and methods focusing on the property of the weights distribution itself. With the first type of methods, the OWA weights can be fitted to empirical data \citep{Yager1996,Beliakov2004}, function of the criteria values \citep{Yager1993,Xu2006} or based on preference relations \citep{Ahn2008}. Although these methods are very interesting and closely related to real application problems, they are usually case-dependent and rely heavily on the quality of data sources. Therefore, in this paper, we will focus on the latter type of methods, the results of which essentially depend on the structure and properties of the OWA weight vector. These methods can be divided into two main categories: the optimization based methods and the function based methods. Optimization based methods usually rely on the concept of \emph{orness} \citep{Yager1988} which measure the \enquote{orlike} level of OWA operators. It can be seen as a measure of risk in decision analysis \citep{Jiang2000,Drobne2009}. The objective of the optimization based methods is basically to maximize the dispersion of the weights (i.e trade-off between criteria) for a given level of \emph{orness} (i.e. risk) \citep{OHagan1988,Yager1994,Filev1995,Filev1998,Fuller2001,Liu2012}. The second class of methods proposes to generate OWA weights using continuous mathematical functions whose shape will determine the OWA weights distribution after a discretization step. The first functions that were proposed relied on linguistic quantifiers to guide the aggregation process \citep{Yager1993,Yager1996}. Different forms of quantifiers and their relationships with \emph{orness} and maximum dispersion has been studied in the past few years \citep{Liu2005,Liu2006,Liu2008a,Liu2008b}. Yager proposed also a particular class of function called \enquote{stress function} that allow greater emphasis to be placed on particular criteria \citep{Yager2007a}. Another interesting class of functions rely on probability theory. Xu proposed in 2005 the use of normal distributions to generate OWA weights \citep{Xu2005}. This idea was taken up by Yager in a paper dealing with centered OWA operator \citep{Yager2007b} and generalized by Sadiq and Tesfamarian to other families of probability density functions (PDF) \citep{Sadiq2007}.  

It is interesting to note that none of these methods allow for a full exploration of the decision strategy space defined by a certain level of risk (\enquote{orlike} level) but also a certain level of trade-off (dispersion) \citep{Jiang2000}. Measuring the impact of the level of risk and trade-off on a decision represents however a crucial step in a decision-making process. It is therefore important to propose methods that automatically generates OWA weights according to a certain level of risk and trade-off allowing for a systematic and rigorous sensitivity analysis assessing the impact of these values on the final output. Therefore, we fundamentally believe that the generation of OWA weights requires a clear understanding of existing relationships between weight distribution, risk and trade-off. To the best of our knowledge, there is no method that allows for an automatic generation of order weights according to a certain level of risk and tradeoff. 

In this article, we tackle this issue by disentangling the relationships between OWA weights, risk and trade-off using truncated distributions. We will first describes the basic concept of OWA operators and formally introduced the concepts of risk and trade-off. We will then explain in details the method that we developed to generate OWA weights according to a certain level of risk and trade-off. In particular, we will focus on the truncated normal distribution and the definition of a continuous parabolic decision-strategy space. Finally, we will explore the influence of the number of criteria on the results.

\section*{OWA operator, risk and trade-off}

Let us consider a collection of $n$ objects $(x_1,...,x_n)$ representing preference values associated with a decision to take on a particular issue. The problem is to aggregate the $n$ objects together in order to obtain an unique object to enable a production decision. The considered objects may have several dimensions or even be weighted but the decision will be always made on an element-by-element basis, so without loss of generality let us consider that the $x_i$ are real numbers, hereafter called criteria.    

To solve this problem, Yager in 1988 introduced a new family of aggregation techniques called ordered weighted averaging (OWA) operators \citep{Yager1988}. Formally, the OWA operator $O_w: \mathbb{R}^n \longrightarrow \mathbb{R}$ associates a set of order weights $w=(w_1,...,w_n)$ such that $\sum_{i=1}^n w_i = 1$ to our collection of $n$ criteria as follows:
\begin{equation}
   O_w(x_1,...,x_n)=\sum_{i=1}^n w_i x_{(i)}
   \label{OWA}
\end{equation}
where $x_{(i)}$ is the $i^{th}$ lowest value in $(x_1,...,x_n)$. The first order weight $w_1$ is therefore assigned to the criterion with the lowest value, the second order weight $w_2$ to the second lowest criterion, etc.. Two dimensions are usually associated with $w$: risk and trade-off. The first dimension characterizes the level of risk in the aggregation process, which positions the values obtained with the aggregation operator $O_w$ on a continuum between the \emph{minimum} and the \emph{maximum} $x_i$ values. The level of risk can be measured from $w$ using the concept of \emph{orness} or \emph{andness} as follows:
\begin{equation}
   \begin{cases}
     andness(w)=\frac{1}{n-1} \sum_{i=1}^n w_i(n-i)\\
	   orness(w)=1-andness(w)
	 \end{cases}		
   \label{ORAND}
\end{equation}
\emph{Orness} and \emph{andness} are two complementary values that reflect an attitude toward risk in decision making. An \emph{andness} value equal to 1 (or \emph{orness} equal to 0) gives full weight to the \emph{minimum} $x_i$ value (intersection, AND operator). It represents a risk-aversion position where the vector of OWA order weights $w$ becomes $(1,0,...,0)$.  On the other extreme, an \emph{orness} equal to 1 (or \emph{andness} equal to 0) gives full weight to the \emph{maximum} $x_i$ value (union, OR operator). It represents the most risk-taking attitude where the vector of OWA order weights $w$ is equal to $(0,0,...,1)$.

\begin{figure}[!ht]
  \centering 
  \includegraphics[width=\linewidth]{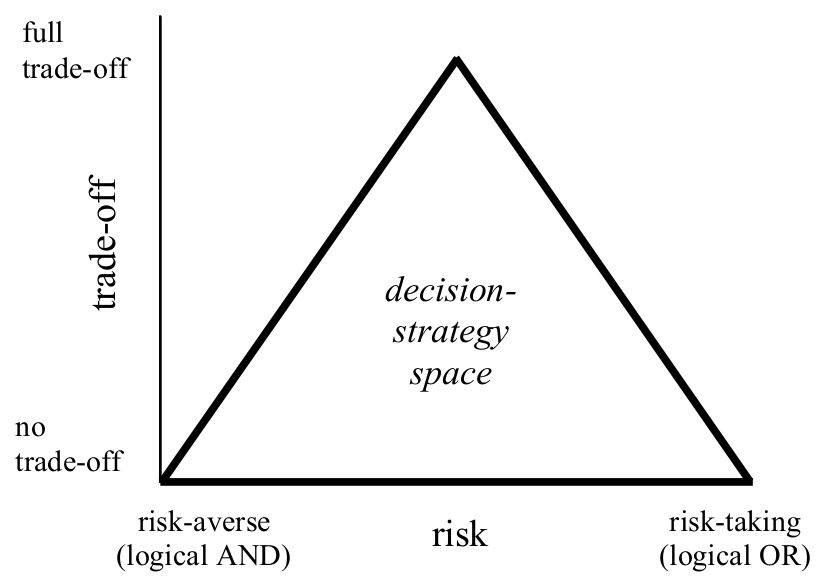}
  \caption{\textbf{Triangular decision-strategy space defined by the dimension of risk and trade-off.} The figure is taken from \citep{Drobne2009}. \label{Fig1}}
\end{figure}

The second dimension represents the level of trade-off between criteria. It can be seen as a measure of dispersion over the OWA order weights. The concept of dispersion in OWA was first introduced by Yager in 1988. Similar to the entropy of Shannon, the measure of dispersion can be computed as follows:
\begin{equation}
    Disp(w)=\frac{-\sum_{i=1}^n w_i\cdot log(w_i)}{log(n)} 		
   \label{E}
\end{equation}
This measure is comprised between 0 and 1, reaching its maximum when $w_i=1/n$. Another measure of tradeoff was proposed by \citep{Jiang2000}, computed as a distance to the uniform distribution,
\begin{equation}
    tradeoff(w)=1-\sqrt{\frac{n\sum_{i=1}^n (w_i-1/n)^2}{n-1}} 		
   \label{tr}
\end{equation}
Various measures could be considered to assess the level of risk and tradeoff associated with a vector of OWA order weights $w$. Without loss of generality, they can be expressed as a couple of values $(\alpha_w,\,\delta_w) \in [0,1]^2$. It is however important to note that these two dimensions are not independent, certain couple of values are inconsistent. It is indeed not possible to obtain a vector $w$ which exhibits at the same time a low or high level of risk and a high level of tradeoff. As shown in Figure \ref{Fig1}, this two dimensional space can be approximated by a triangle.

\begin{figure}[!ht]
  \centering 
  \includegraphics[width=7cm]{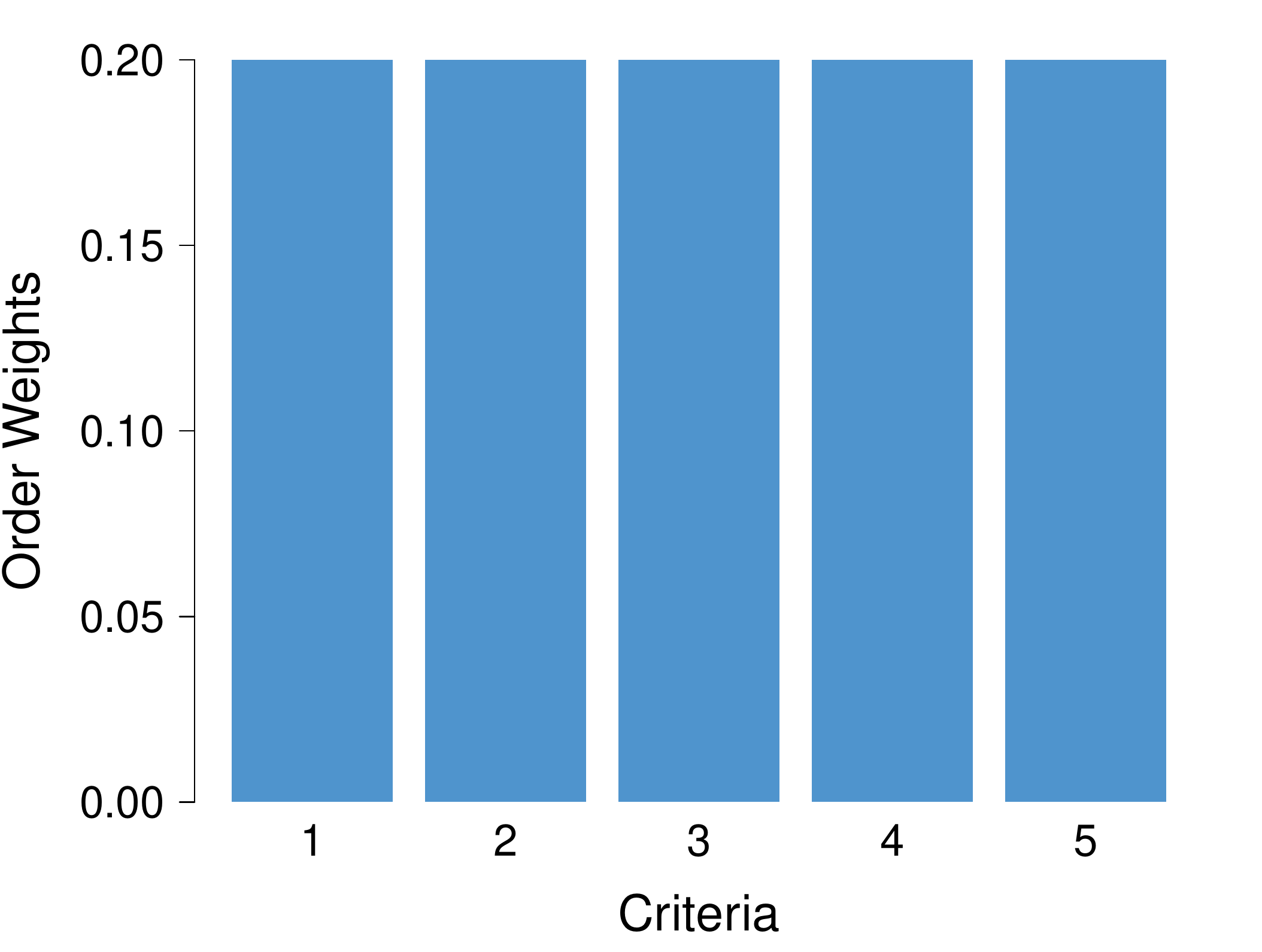}
  \caption{\textbf{Distribution of order weights for five criteria with average level of risk ($\alpha_w=0.5$) and full trade-off ($\delta_w=1$).} \label{Fig2}}
\end{figure}

The three vertices of the triangle represents the three main OWA operators: minimum $(\alpha_w=0,\,\delta_w=0)$, maximum $(\alpha_w=1,\,\delta_w=0)$ and average $(\alpha_w=0.5,\,\delta_w=1)$. The average operator corresponds to a weighted linear combination (WLC) of criteria with a middle level of risk ($\alpha_w=0.5$) and a full trade-off ($\delta_w=1$). An example of such distribution of order weights for 5 criteria is presented in Figure \ref{Fig2}. Except for some trivial cases, like the ones described above, it is not formally established, in practice, how to generate order weights according to a certain level of risk and trade-off. 

\section*{Generation of OWA weights using truncated distributions}

\begin{figure}[!ht]
  \centering 
  \includegraphics[width=7cm]{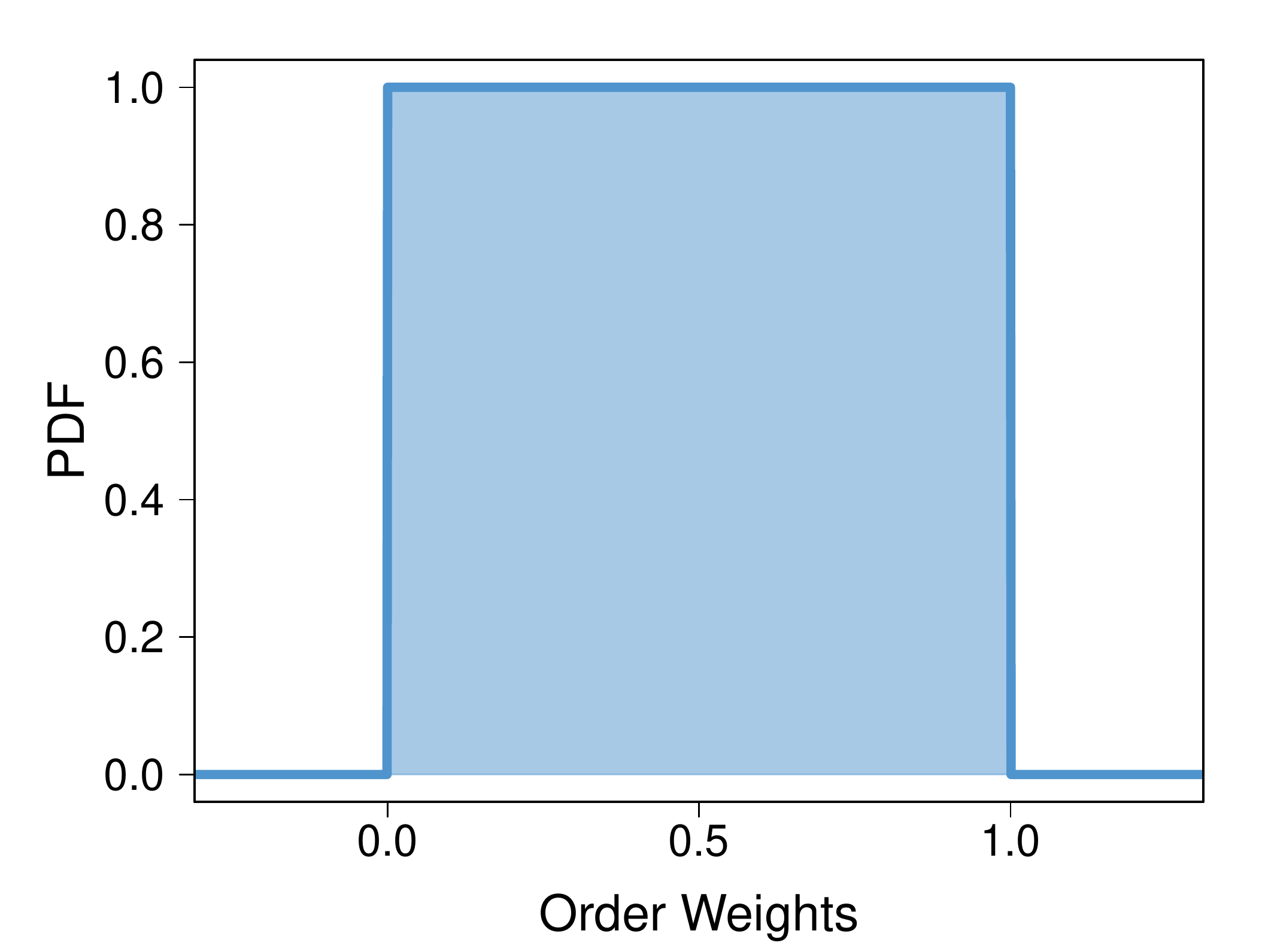}
  \caption{\textbf{Continuous distribution of order weights with average level of risk ($\alpha_w=0.5$) and full trade-off ($\delta_w=1$).} \label{Fig3}}
\end{figure}

The generation of OWA weights generation can be investigated through the existing relationships between order weights, risk and tradeoff. Indeed, a formal description of the relationships between these concepts will permit us to not only better understand the link between risk and tradeoff (i.e. decision-strategy space), but to propose a way to generate automatically OWA order weights from any risk and tradeoff values in this triangular decision-strategy space. Here, we tackle this problem by proposing an approach relying on probability density function (PDF) to generate a vector of order weights $w$ according to a couple of risk and trade-off values $(\alpha_w,\,\delta_w)$ comprised between 0 and 1. Indeed, as already pointed out in several studies, OWA order weights can be viewed as a discrete probability density function represented by a vector of real numbers comprised between 0 and 1 that sum to 1 \citep{Xu2005,Sadiq2007}. Therefore, an OWA weights distribution can be modeled asymptotically (when $n\rightarrow \infty$) as a continuous probability density distribution $f_w$. An example is given in Figure \ref{Fig3}, in which the continuous version of the distribution of order weights presented in Figure \ref{Fig2} is displayed. 

\begin{figure*}
  \centering 
  \includegraphics[width=\linewidth]{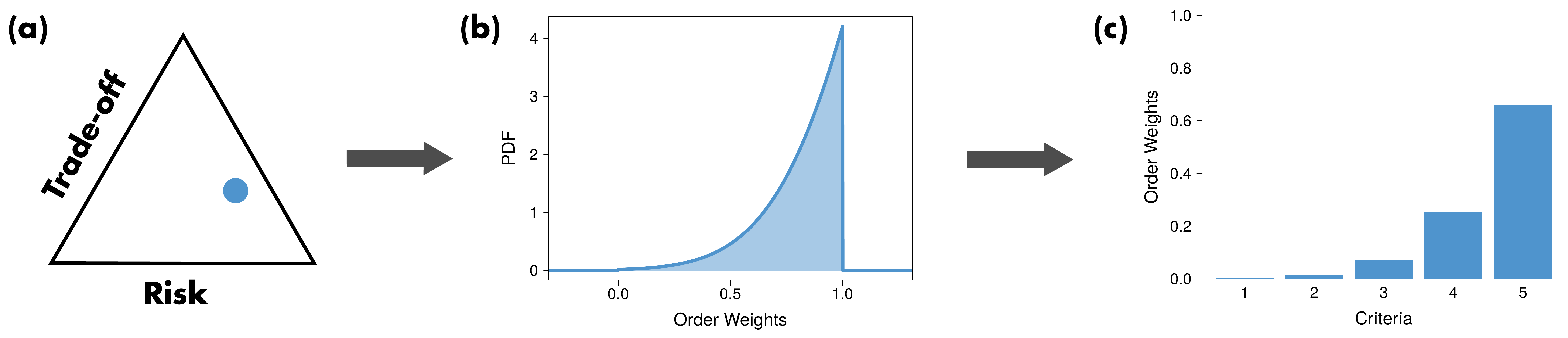}
  \caption{\textbf{Automatic generation of OWA weights according to a certain level of risk and trade-off based on truncated probability distribution.} (a) Choose a certain level of risk and trade-off. (b) Generation of a continuous OWA order weights distribution according to a certain level of risk and trade-off. The continuous distribution take the the form of a probability density function of a truncated probability distribution. (c) Discretization of the PDF in order to obtain $n$ order weights. \label{Fig4}}
\end{figure*} 

It is important to note that, unlike other approaches, the domain of $f_w$ is bounded and, for simplicity purposes, is set to $[0,1]$. This choice of a bounded domain for $f_w$ offers several advantages. The main one is to consider truncated distributions characterized by bounded averages and standard deviations. The average $\mu_w$ is comprised between 0 and 1, and the standard deviation $\sigma_w$ is upper-bounded by $\frac{1}{2\sqrt{3}}$, standard deviation of a uniform distribution $\mathcal{U}_{[0,1]}$. Average and standard deviation represent the two first moments of a PDF, and these two statistics have several properties in common with the risk and trade-off measures associated with OWA order weights. As described in Equation \ref{ORAND}, \emph{andness} and \emph{orness} are governed by the skewness of $w$ \citep{Jiang2000}. In our case, the level of asymmetry of $f_w$ depends on $\mu_w$, positive skew when $\mu_w < 0.5$ and negative skew when $\mu_w > 0.5$. In the same way, Equation \ref{E} or \ref{tr} shows that the level of tradeoff is controlled by the degree of dispersion in the order weights \citep{Jiang2000}. In a probability distribution, the degree of dispersion is represented by the standard deviation $\sigma_w$. We also observe the same relationship between $\mu_w$ and $\sigma_w$ and $\alpha_w$ and $\delta_w$. In both cases, high levels of asymmetry are incompatible with a high level of dispersion and the maximum dispersion can be reached when the average equal the median (Figure \ref{Fig4}). The main difference between the two couples of metrics is that the upper-bound for dispersion is different. Thus, we define a relationship between risk and trade-off and average and standard deviation as follows,
\begin{equation}
    \begin{cases}
      \alpha_w=\mu_w \\
      \delta_w=2\sqrt{3} \, \sigma_w
    \end{cases}
   \label{ad}
\end{equation}
Hence, any truncated distribution restricted on the domain $[0,1]$ satisfying Equation \ref{ad} can be seen as a continuous distribution of OWA order weights with level of risk and trade-off equal to $\alpha_w$ and $\delta_w$, respectively. Using truncated distribution offers also a simple way to describe how the OWA order weights are generated from $f_w$. Equation \ref{wi} is indeed required to discretize $f_w$ and obtain a vector of $n$ OWA order weights.
\begin{equation}
   w_i=\frac{f_w \left(\frac{i-1}{n-1}\right)}{\sum_{k=1}^n f_w \left(\frac{k-1}{n-1}\right)}\,\,,\,\,\forall i \in |[1,n]|
   \label{wi}
\end{equation}
Nevertheless, it is important to ensure that the denominator in Equation \ref{wi} is not equal to 0. In particular, when $\delta_w=\sigma_w=0$, $f_w$ is represented by a generalized probability density function $f_w(x)=\delta(x-\alpha_w)$ using the Dirac delta function. In this particular case, the weight with normalized index $\frac{i-1}{n-1}$ nearest to $\alpha_w$ is equal to 1 and all the others are set to 0. 

Hence, we showed that truncated distributions can be used to generate automatically OWA weights from any risk and tradeoff values in this triangular decision-strategy space (Figure \ref{Fig4}).

\section*{Continuous parabolic decision-strategy space}

In practice, not all families of probability distribution are suitable to generate meaningful OWA weights. In this study, we focused on the truncated normal distribution $\mathcal{N}_{[0,1]}(\mu_w,\sigma_w)$ restricted on the domain $[0,1]$. $\mu_w$ and $\sigma_w$  represent the average and standard deviation computed after truncation, respectively. As explained in the previous section, for a certain level of risk $\alpha_w \in [0,1]$, and trade-off $\delta_w \in [0,1]$, one can model a continuous OWA order weight distribution $f_w$ using a truncated normal distribution $\mathcal{N}_{[0,1]}(\mu_w,\sigma_w)$ satisfying Equation \ref{ad}. 

However, an important issue that we need to address concerns the generation of the truncated normal distribution itself. In fact, any normal distribution with average $\mu$ and standard deviation $\sigma$ lying within the interval $[0,1]$ is a truncated normal distributions of average $\mu_w$ and standard deviation $\sigma_w$. Its probability density function is given by
\begin{equation}
   f(x)=\frac{\phi\left(\frac{x-\mu}{\sigma}\right)}{\sigma\left( \Phi\left(\frac{1-\mu}{\sigma}\right) - \Phi\left(-\frac{\mu}{\sigma}\right) \right)} 
   \label{f}
\end{equation}
when $x \in [0,1]$ and 0 otherwise. $\phi$ is the PDF of the standard normal distribution and $\Phi$ is its cumulative distribution function. The mean and standard deviation of the truncated distribution on the domain $[0,1]$ can be computed with the formula displayed in Equation \ref{muw} and \ref{sdw}, respectively.
\begin{equation}
   \mu_w=\mu + \sigma \frac{\phi\left(-\frac{\mu}{\sigma}\right) - \phi\left(\frac{1-\mu}{\sigma}\right)}{\Phi\left(\frac{1-\mu}{\sigma}\right) - \Phi\left(-\frac{\mu}{\sigma}\right)} 
   \label{muw}
\end{equation}
\begin{equation}
   \hspace{-1cm}\sigma_w=\sigma\sqrt{1+\frac{\frac{1-\mu}{\sigma}\phi\left(\frac{1-\mu}{\sigma}\right)+\frac{\mu}{\sigma}\phi\left(-\frac{\mu}{\sigma}\right)}{\Phi\left(\frac{1-\mu}{\sigma}\right) - \Phi\left(-\frac{\mu}{\sigma}\right)}-\left( \frac{\phi\left(-\frac{\mu}{\sigma}\right) - \phi\left(\frac{1-\mu}{\sigma}\right)}{\Phi\left(-\frac{\mu}{\sigma}\right) - \Phi\left(\frac{1-\mu}{\sigma}\right)}\right)^2} 
   \label{sdw}
\end{equation}
Consequently, to generate a truncated normal distribution $\mathcal{N}_{[0,1]}(\mu_w,\sigma_w)$, we first need to identify the normal distribution $\mathcal{N}(\mu,\sigma)$ whose restriction on the domain $[0,1]$ will give us an average of $\mu_w$ and a standard deviation equal to $\sigma_w$. Although it exists a clear relationship $(\mu,\sigma) \rightarrow (\mu_w,\sigma_w)$ allowing for the generation of a truncated normal distribution $f_w$ from any normal distribution of parameter $\mu$ and $\sigma$, the opposite is not true. Indeed, as previously explained, the more $\mu_w$ deviates from $0.5$, the smaller the maximal value that $\sigma_w$ can take, thus excluding a wide range of combinations.

To overcome this issue, we have decided here to favor a numerical approach. Let us consider a couple of risk and trade-off values $(\alpha_w,\delta_w)$ comprised between 0 and 1. The associated average and standard deviation $(\mu_w,\sigma_w)$ can be easily derived from Equation \ref{ad}. We then need to identify the couple of values $(\mu,\sigma)$ whose \enquote{truncated} average and standard deviation $({\hat{\mu}_w},{\hat{\sigma}_w})$ on the domain $[0,1]$ are the closest from $(\mu_w,\sigma_w)$. To do so, we used an optimization algorithm \citep{Nelder1965} which seeks to find the optimal combination of $\mu$ and $\sigma$ values minimizing the quadratic distance $d$ between the expected $(\mu_w,\sigma_w)$ values and the simulated ones (Equation \ref{L2}).
\begin{equation}
   d=\sqrt{(\mu_w-{\hat{\mu}_w})^2+(\sigma_w-{\hat{\sigma}_w})^2} 
   \label{L2}
\end{equation}
At the end of the process, we obtain a candidate probability density function $f_w$, the validity of which is evaluated with $d$. 

\begin{figure}[!ht]
  \centering 
  \includegraphics[width=8cm]{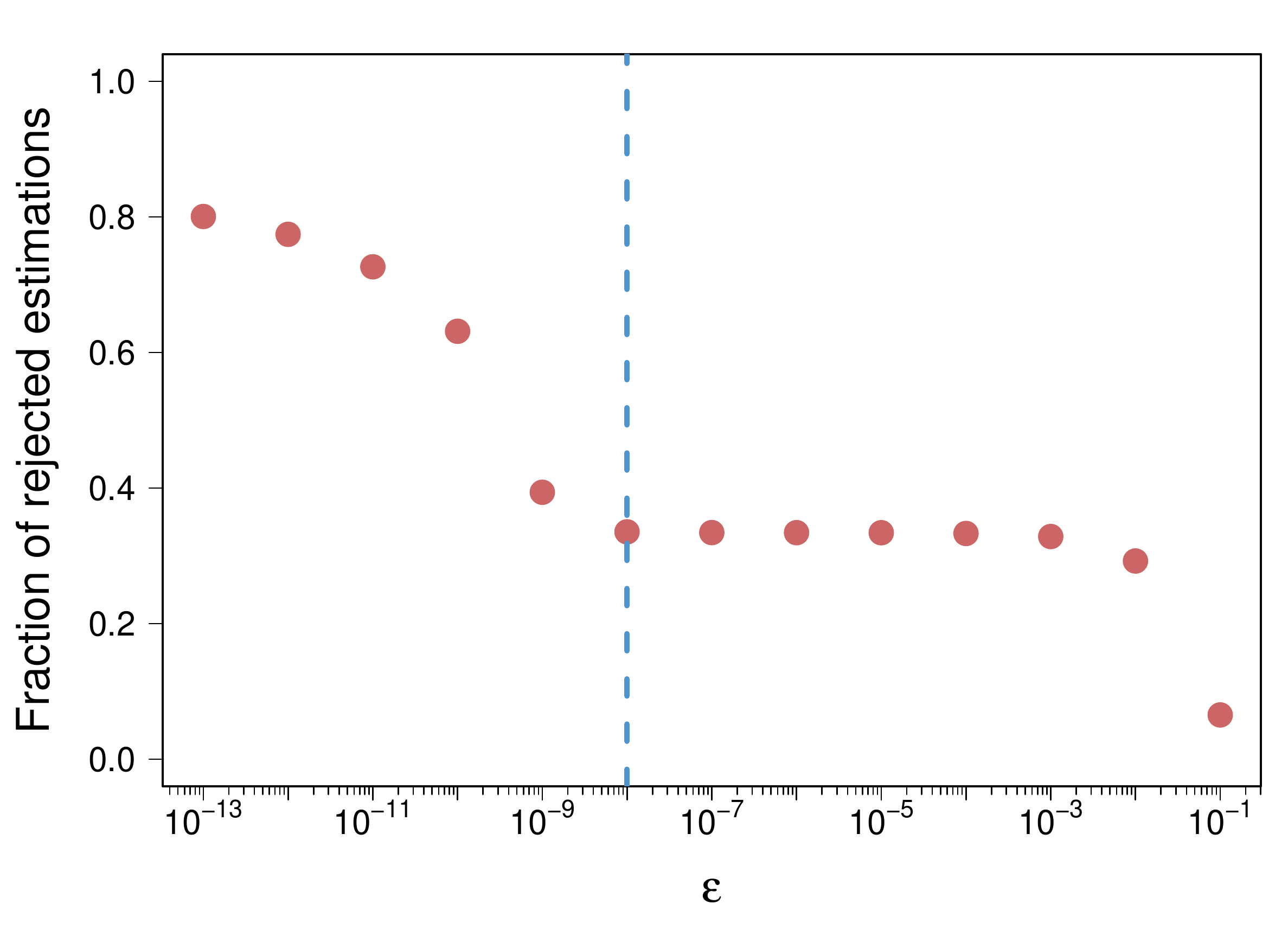}
  \caption{\textbf{Evolution of the fraction of rejected estimations as a function of $\epsilon$.} The blue dashed line represent the optimal $\epsilon$ value.\label{Fig5}}
\end{figure}

\begin{figure*}
  \centering 
  \includegraphics[width=13cm]{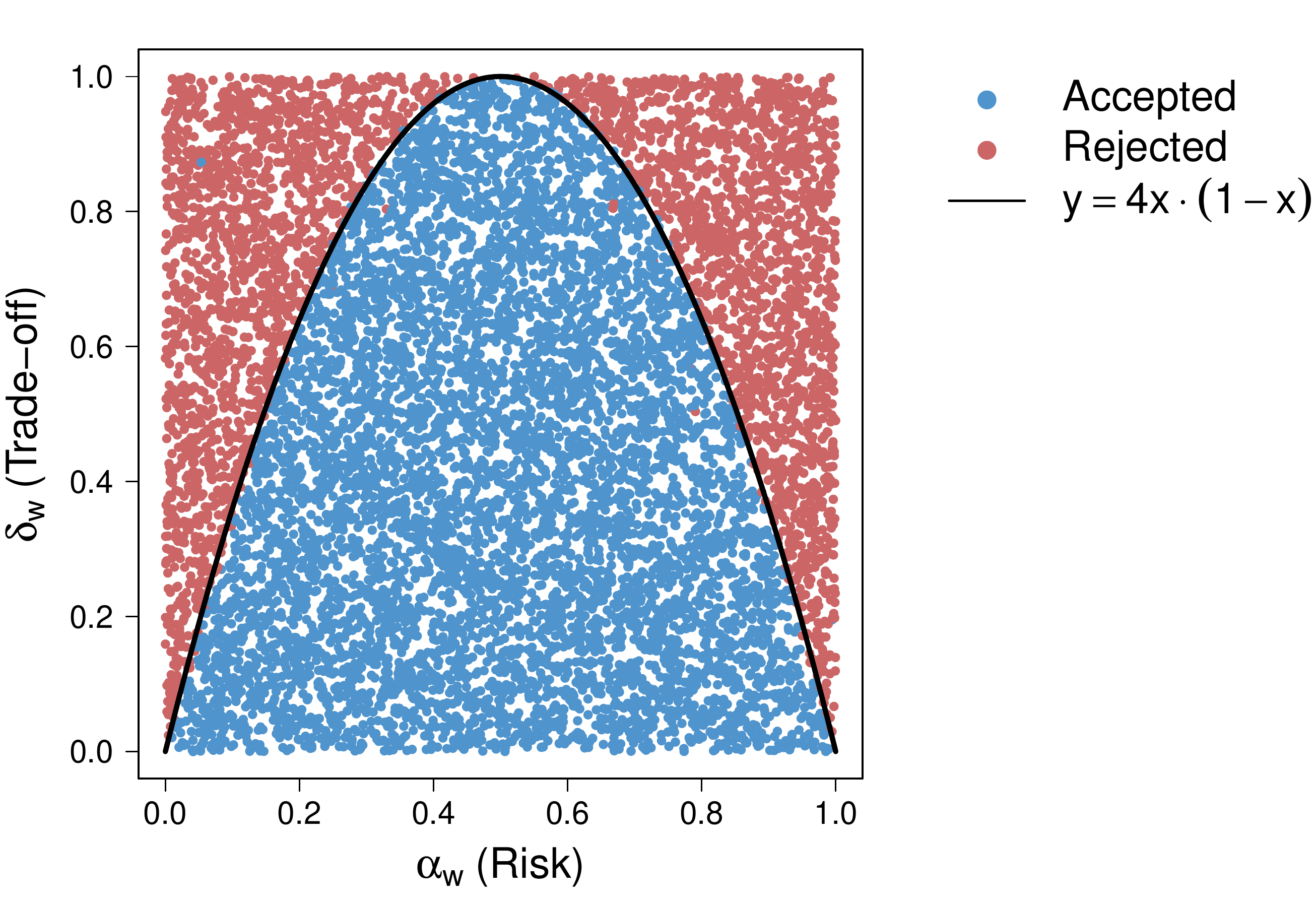}
  \caption{\textbf{Representation of the continuous parabolic decision-strategy space defined by the dimension of risk and trade-off based on truncated normal distribution.} The blue dots represent the couple of risk and trade-off value for which $d<\epsilon$. The red dots represent the couple of risk and trade-off value for which $d\geq\epsilon$. The black line represents the frontier between the two regions approximated with a parabola.  \label{Fig6}}
\end{figure*}

We assume that if $d$ is higher than a predetermined threshold $\epsilon$, then no suitable PDF can be found with these risk and trade-off values. A simple way to calibrate the threshold value $\epsilon$ is to investigate its impact on the fraction of rejected estimations such as $d\geq\epsilon$. For that purpose, we drawn at random $10,000$ couple of risk and trade-off values using a latin hypercube sampling. Then, we applied the process described above to each element of our sample in order to quantify the distance $d$ associated with each of them. As it can be observed in Figure \ref{Fig5}, the fraction of rejected estimations is high for very small values of $\epsilon$. Then, it starts to decrease when $\epsilon$ increases, until it reaches a plateau between $10^{-8}$ and $10^{-3}$. This plateau represents a gap between rejected and accepted estimations. Figure \ref{Fig6} shows the two dimensional representation of the rejected and accepted estimations with $\epsilon=10^{-8}$ (beginning of the plateau) as a function of the level of risk and trade-off. It is interesting to observe that the frontier between rejected and accepted estimations coincides with a parabola whose equation is $y=4x(1-x)$. It is therefore possible to formally describe the relationship between risk and trade-off and define a continuous parabolic decision-strategy space to automatically generates OWA weights.   

\begin{figure*}
  \centering 
  \includegraphics[width=16cm]{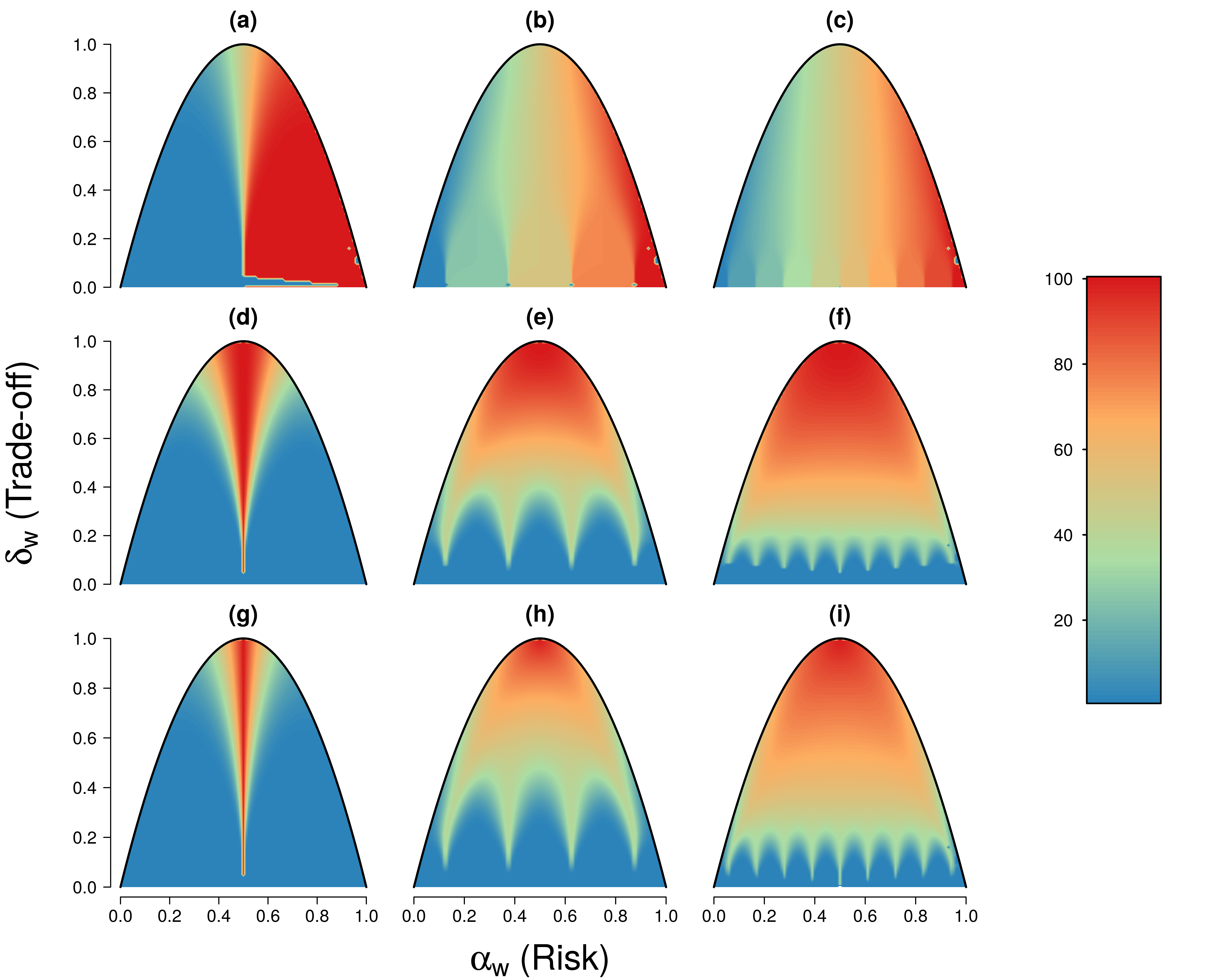}
  \caption{\textbf{Influence of the number of criteria on the OWA weights' properties generated with the truncated normal distribution for a given level of risk and trade-off.} From the left to the right, the results obtained with two criteria ((a), (d) and (g)), five criteria ((b), (e) and (h)) and ten criteria ((c), (f) and (i)) are displayed. From the top to the bottom, the properties of the OWA weight distribution are measured with the $orness$ as defined in Equation \ref{ORAND} ((a)-(c)), the dispersion as defined in Equation \ref{E} ((d)-(f)) and trade-off as defined in Equation \ref{tr} ((g)-(i)). \label{Fig7}}
\end{figure*}

\section*{Influence of the number of criteria}

In this section, we investigate the influence of the number of criteria, $n$, on the properties of the weight distribution generated with the truncated normal distribution for a given level of risk and trade-off. It is indeed important to keep in mind that the discretization step (Equation \ref{wi}) have a non-negligible effect on the shape of the weight distribution, particularly when the number of criteria is small.  To do so, we relied on three well-known metrics described in the first part of the paper to evaluate the properties of the OWA weight distribution: the $orness$, the $Disp$ and the $tradeoff$ as defined in Equation \ref{ORAND}, \ref{E} and \ref{tr}, respectively. The evolution of these metrics as a function of $\alpha_w$ and $\delta_w$ for different values of $n$ (2, 5 and 10) are displayed in Figure \ref{Fig7}. It can be observed that the weight distributions are less sensitive to variations in $\alpha_w$ and $\delta_w$ values when the number of criteria decreases. This is particularly true for small $\delta_w$ values. Indeed, when $n$ decreases, the level of trade-off required to find a compromise between at least two criteria ($w_i<1$, $\forall i$) becomes higher. This phenomenon is amplified by the level of risk, $\alpha_n$, the value of which may reflect a certain willingness to compromise between two consective criteria $i$ and $i+1$ according to its position in the range bin $[\frac{i-1}{n-1},\frac{i-1}{n-1}]$. Although the discretization step has very strong effect on the weight generation for very small numbers of criteria, $n=2$ or $n=3$, it becomes less and less significant as $n$ increases. Besides, it is also worth noting that the dispersion associated with $w$ obtained for a given value of $\alpha_w$ and $\delta_w$ tend to increase when $n$ increases. This is a property of the dispersion as defined in Equation \ref{E} and \ref{tr}, which values tend to be closer to $1$ as $n\rightarrow \infty$ \citep{Malczewski2006}.

\section*{Discussion}

In summary, we introduced a new method to determine OWA weights. The method is based on probability density function of truncated distributions. It allows for an automatic generation of OWA weights according to a certain level of risk and trade-off based on the two first moments of the probability distribution. The main advantage of our method is to provide a rigourous framework for a full exploration of the decision-strategy space (Figure \ref{Fig4}). Moreover, the proposed method enables to conduct systematic sensitivity and uncertainty analysis on multi-criteria decision analysis and makes it possible to better assess the impact of the level of risk and trade-off on the final decision.

In this work, we focused on the truncated normal distribution and unveiled the relationship between risk and trade-off by identifing a well-defined continuous parabolic decision-strategy space. We have then examined the effect of dizcretization step on the weight distribution. Unsurprisingly, we showed that the sensitivity to variations in risk and trade-off values with the number of criteria. An inevitable direction for further studies will be to extent this analyis to other families of PDFs.   

We are especially interested in proposing a method accessible to non-specialist  policymakers. This is why we decided to focus on OWA weights’ generation through probability density functions. They have indeed the clear advantage of simplicity while relying on well-established statistical properties. Finally, the software package to generate OWA using the approach described in the paper along with a interactive web application for visualizing the results can be downloaded from \url{https://github.com/maximelenormand/OWA-weights-generator}.

\section*{Acknowledgements}

We thank Vahid Rahdari for useful discussions. 

\bibliographystyle{unsrt}
\bibliography{OWG}
 
\end{document}